\documentclass[11pt]{article}

\usepackage[final]{acl}

\usepackage{times}
\usepackage{latexsym}

\usepackage[T1]{fontenc}
\usepackage[utf8]{inputenc}

\usepackage{microtype}

\usepackage{inconsolata}

\usepackage{graphicx}
\usepackage{booktabs}
\usepackage{makecell}
\usepackage{tabularx}
\usepackage{adjustbox}
\usepackage{multirow}
\usepackage{float}
\usepackage{amsmath}

\usepackage{listings}
\usepackage{xcolor}

\title{GeoBuildBench: A Benchmark for Interactive and Executable Geometry Construction from Natural Language}

\author{
 \textbf{Jinwoong Kim \textsuperscript{1,2}},
 \textbf{Rui Yang \textsuperscript{1}},
 \textbf{Huishuai Zhang \textsuperscript{1,2}},
\\
 \textsuperscript{1}Peking University
 \textsuperscript{2}Wangxuan Institute of Computer Technology,
\\
\texttt{\{jinwoongkim, ypyangrui, zhanghuishuai\}@pku.edu.cn}
}

\begin{document}
\maketitle
\begin{abstract}
We introduce \textbf{GeoBuildBench}, a benchmark designed to evaluate whether large language models and multimodal agents can ground informal natural-language plane geometry problems into executable geometric constructions. Unlike existing geometry benchmarks that focus on answer correctness or static diagram interpretation, GeoBuildBench treats geometry diagram as an interactive construction task: given a textual problem, an agent must generate a domain-specific language (DSL) program to produce a diagram satisfying explicitly specified geometric objects and verifiable constraints. The benchmark features 489 Chinese textbook-style problems, curated through automated filtering and human validation to ensure text-complete, constructible problem specifications. We evaluate several state-of-the-art multimodal models in a bounded iterative setting and show that, despite reasonable success rates, models frequently exhibit structural hallucinations, missing objects, and failures to satisfy geometric constraints, with limited ability to exploit visual and constraint-based feedback for self-correction. These results highlight geometry construction as a rigorous testbed for grounded, executable reasoning beyond textual or visual plausibility. Our benchmark and
code are released at \url{https://github.com/ooongs/GeoBuildBench}. 
% \rui{Replace with anonymous repo link?}
\end{abstract}

%%%%%%%%%%%%%%%%%%%%%%%%%%%%%%%%%%%%%%%%%%%%%%%%%%%%%%%%%%%%

\section{Introduction}
\label{sec:introduction}
Large language models (LLMs) and multimodal LLMs (MLLMs) have shown strong performance on geometry-focused benchmarks such as GeoQA and Geometry3K, which evaluate multimodal reasoning over textbook-style geometry problems and diagrams \cite{chen-etal-2021-geoqa,lu2021intergpsinterpretablegeometryproblem}. In parallel, neuro-symbolic systems such as AlphaGeometry achieve near–Olympiad-level performance by combining language models with formal deduction engines \cite{trinh2024alphageometry}. Despite this progress, existing approaches largely assume formalized representations or given diagrams, leaving open a fundamental question: can these models transform informal, natural-language descriptions into consistent, executable geometric structures?

Geometry provides a uniquely stringent testbed for grounded reasoning. Textbook problems describe points, lines, and geometric relations in natural language, implicitly requiring the construction of a consistent diagram before symbolic reasoning. Errors in this process are rarely ambiguous: incorrect or missing constructions typically lead to impossible or visibly invalid diagrams, making geometry well suited for studying structural hallucination.

Prior work has often treated diagram construction as a one-shot preprocessing step, using relation-extraction to map text into solver-specific representations~\cite{gan2018formalization} or numerical optimization~\cite{krueger2021gmbl} to produce coordinates. More recent text-to-diagram methods emphasize coordinate accuracy and formal constraints \cite{hu2023text2diagram,wang2025magicgeo,cheng2025geouni}, but their evaluations rely on visual similarity or parser-dependent checks, and do not assess whether agents can incrementally build, inspect, and repair geometric structures under explicit semantic constraints. Although agentic frameworks like ReAct and Reflexion~\cite{yao2023react,shinn2024reflexion} have shown promise in interleaving reasoning with actions, they have primarily been tested in textual environments. Geometry differs fundamentally: every action must satisfy strict spatial constraints, and failures, such as intersecting parallel lines or failing a tangency requirement, are immediately observable through the environment.

\begin{figure*}

  \centerline{\includegraphics[width=0.8\textwidth]{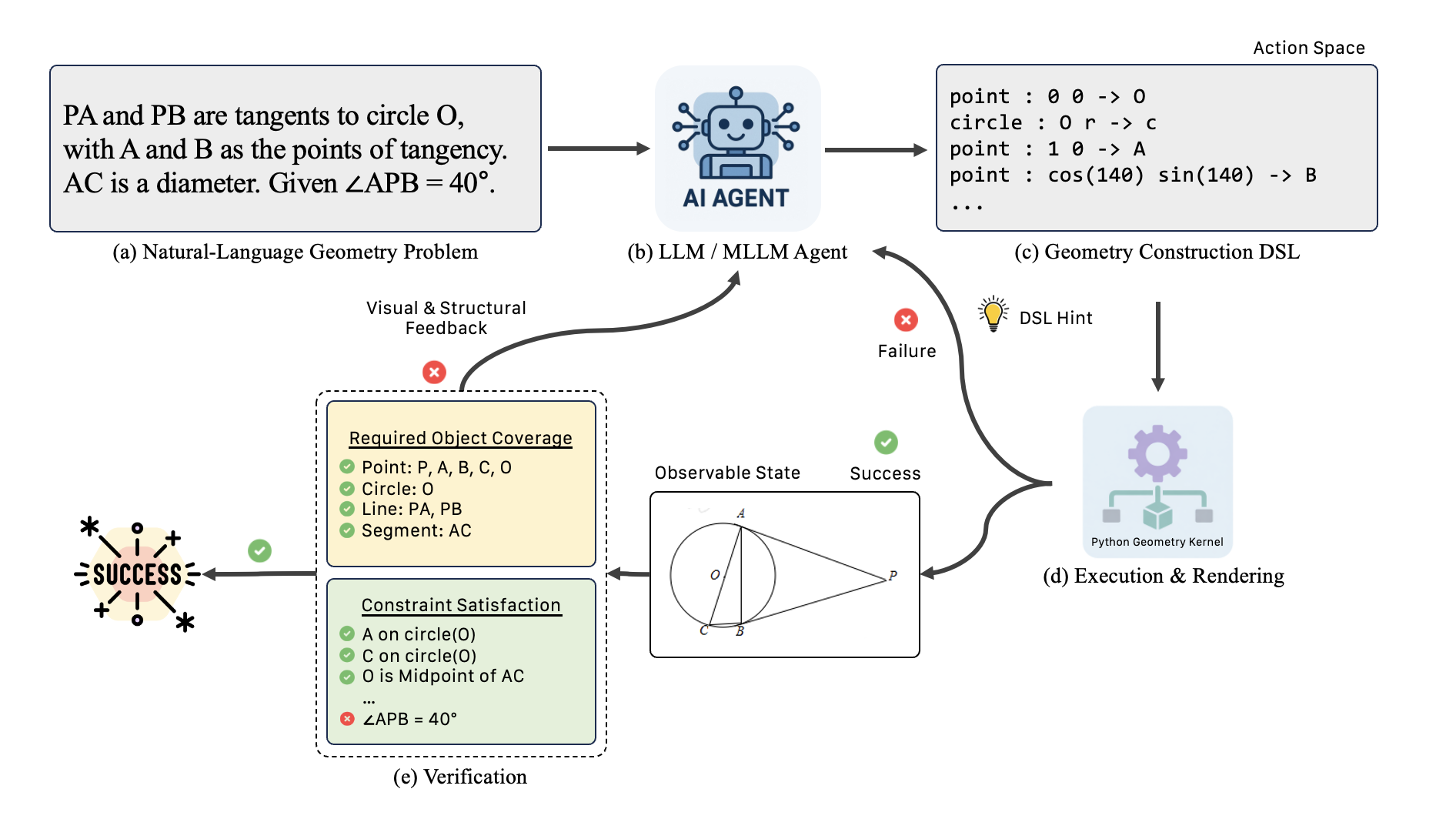}}
  \caption{
Overview of the GeoBuildBench environment.
An agent translates a natural-language geometry problem into executable DSL programs,
executes them to obtain rendered diagrams, and verifies the result against required object coverage,
construction validity, and geometric constraints.
Violation feedback is returned to the agent for iterative repair, while the interaction terminates
once all required objects are present and all constraints are satisfied.
}
  \label{fig:overview}
\end{figure*}

To address this gap, we introduce \textbf{GeoBuildBench}, a benchmark and interactive environment for evaluating agents that translate natural-language plane geometry problems into executable geometric constructions. We define a compact, execution-oriented geometry Domain Specific Language (DSL), together with a Python interpreter and Matplotlib renderer. Instead of evaluating against a single ``gold'' diagram, for each problem GeoBuildBench specifies a set of required objects and explicit verifiable geometric constraints. Agents iteratively generate DSL programs, execute them, and use visual and constraint-based feedback to correct errors, forming a closed agent–environment loop. Figure~\ref{fig:overview} illustrates this agent--environment loop, including the geometry DSL
action space, execution and rendering, and constraint-based verification.

This setting enables us to evaluate models beyond textual correctness or visual plausibility. We measure (i) \emph{executability} of constructions, (ii) \emph{object coverage}, (iii) \emph{geometric constraint satisfaction}, e.g., parallelism and tangency, and (iv) \emph{structural hallucinations}, such as infeasible constructions or references to undefined objects.

Our contributions are threefold:
\begin{enumerate}
    \item \textbf{An Interative Environment:} An executable geometry construction environment with a minimal DSL and deterministic rendering for grounded agent interaction.
    \item \textbf{The GeoBuildBench Dataset:} A benchmark of 489 Chinese textbook-style geometry problems, annotated with constraints to support evaluation without assuming a single gold construction.
    \item \textbf{Empirical Analysis of MLLMs:} An evaluation of state-of-the-art models revealing that current MLLMs struggle with structural correctness and iterative self-repair despite their strong performance on existing geometry QA benchmarks.
\end{enumerate}

%%%%%%%%%%%%%%%%%%%%%%%%%%%%%%%%%%%%%%%%%%%%%%%%%%%%%%%%%%%%

\section{Related Work}
\label{sec:related-work}

\subsection{Geometry Problem Solving Benchmarks and Formal Representations}
\label{subsec:geometry-solving-benchmarks}
Geometry problem solving has long been studied as a multimodal reasoning task over natural-language descriptions, diagrams, and geometric rules. Early systems such as GEOS combine text parsing and diagram interpretation to answer standardized geometry questions \cite{seo-etal-2015-solving}. More recent benchmarks significantly scale up supervision and emphasize structured intermediate representations. GeoQA introduces a large collection of geometry problems paired with executable programs, enabling interpretable multimodal reasoning \cite{chen-etal-2021-geoqa}. Geometry3K further provides densely annotated geometry problems and supports symbolic reasoning through the Inter-GPS framework, which translates textual and diagrammatic inputs into a formal geometry language for theorem-guided inference \cite{lu2021intergpsinterpretablegeometryproblem}. In parallel, prior work has explored formalizing natural-language geometry problems by extracting relations and mapping them into solver-compatible representations \cite{gan2018formalization}. While these datasets and systems advance geometry question answering and symbolic reasoning, they primarily focus on predicting answers or formal programs, and do not directly evaluate whether a model can construct a geometrically consistent diagram from natural language under strict executability and constraint satisfaction.

\subsection{Text-to-Diagram Generation and Constraint-Based Geometry Construction}
\label{subsec:text-to-diagram-gen}
A complementary line of work addresses diagram construction from textual descriptions. Geometry Model Builder (GMB) and its domain-specific language GMBL represent constructions and constraints explicitly, generating diagrams via numerical optimization over geometric constraints \cite{krueger2021gmbl}. Hu and Zhong propose a training-free text-to-diagram method that optimizes point coordinates to produce accurate textbook-style diagrams \cite{hu2023text2diagram}. More recently, MagicGeo and MagicGeoBench combine LLM-based text formalization with a formal geometry solver to generate constraint-consistent diagrams \cite{wang2025magicgeo}. However, their evaluation relies primarily on CLIP-based visual similarity, which measures visual-textual alignment but does not verify whether geometric relations such as parallelism, perpendicularity, or incidence are semantically satisfied. GeoUni explores a unified neural model for diagram generation and geometric reasoning \cite{cheng2025geouni}, but its evaluation pipeline depends on a geometry parser with limited relational coverage, making it difficult to detect violations of unrecognized constraints. In contrast to these approaches, our work focuses on evaluating geometry construction as an executable, constraint-checked process, and on analyzing whether general-purpose LLM-based agents can iteratively build, inspect, and repair geometric structures from natural language.

%%%%%%%%%%%%%%%%%%%%%%%%%%%%%%%%%%%%%%%%%%%%%%%%%%%%%%%%%%%%

\section{GeoBuildBench Environment}
\label{sec:environment}

\subsection{Task Definition}
\label{subsec:task-definition}
GeoBuildBench evaluates an agent’s ability to ground a natural-language plane geometry problem
into an executable geometric construction.
Each task consists of three components:
(i) a natural-language problem description,
(ii) a set of \emph{required geometric objects} that must appear in the final construction, and
(iii) a list of \emph{verification conditions} encoding the geometric semantics of the problem.

The agent’s goal is to generate a construction program that, when executed,
produces a geometric configuration satisfying all required objects and verification conditions.
Importantly, GeoBuildBench does not assume a single gold construction or canonical diagram. 
Multiple constructions are considered valid as long as they are executable and semantically accurate.
This design explicitly encourages the use of \emph{auxiliary constructions} such as additional  helper points,
lines, or circles introduced to satisfy distance or incidence constraints, mirroring standard
human geometric practice.

A task is considered successful if the generated construction 
(1) executes without runtime errors,
(2) contains all required geometric objects in the final state, and
(3) satisfies all verification conditions.
These criteria define a binary notion of task success, while partial metrics can be computed
to analyze intermediate failure modes.
Tasks terminate either upon success or when a predefined interaction budget is exceeded.

We do not assume that the verification conditions exhaustively cover all possible
geometric relations that may appear in natural-language problems.
Instead, each task is specified using a finite and predefined set of condition
types that capture common, executable geometric semantics and admit reliable
numerical verification.

\subsection{Geometry Construction DSL}
\label{subsec:geometry-consgtruction-dsl}
To support precise and deterministic evaluation, GeoBuildBench introduces a compact,
execution-oriented domain-specific language (DSL) for plane geometry construction.
The DSL defines the agent’s action space and is designed to be minimal, explicit, and fully executable.

The DSL provides primitives for constructing geometric entities
(e.g., \texttt{point}, \texttt{line}, \texttt{segment}, \texttt{circle})
and operators for common geometric constructions
(e.g., intersections, parallel and perpendicular lines, midpoints, rotations).
Each command deterministically updates the geometric state and returns named objects that may be
referenced by subsequent commands.
Crucially, the DSL is a \emph{construction language} rather than a declarative constraint language:
relations such as parallelism, perpendicularity, or angle values cannot be asserted directly,
but must be realized through explicit constructions.

The complete DSL syntax, supported commands, and execution semantics are specified in
Appendix~\ref{app:geodsl}.
In this section, we focus on the DSL’s role as an executable action interface rather than
enumerating its full grammar.

\subsection{Execution and Verification}
\label{subsec:exec-veri}
The generated DSL programs are executed by a geometry kernel that maintains an explicit geometric state,
including object coordinates and structural relations.
Execution failures, such as referencing undefined objects or attempting geometrically infeasible
operations (e.g., intersecting parallel lines), are explicitly detected and reported. 

After execution, a verification module assesses the result across three dimensions:
\begin{itemize}
    \item \textbf{Executability} checks whether the program runs to completion without errors.
    \item \textbf{Required Object Coverage} verifies that all geometric objects specified in the task definition are present in the final construction, regardless of any auxiliary objects introduced.
    \item \textbf{Constraint Satisfaction} checks if all verifiable conditions like incidence, parallelism, metric constraints, and angular requirements are met.
\end{itemize}

Violations of these checks reveal different forms of \emph{structural hallucination},
such as references to non-existent objects, geometrically infeasible constructions,
or failures to satisfy required relations.
Formal definitions of all verification condition types and their numerical tolerances are provided
in Appendix~\ref{app:verification}.
This explicit verification enables semantic evaluation beyond textual correctness
or visual similarity, without reliance on proxy image-based metrics.

\subsection{Agent--Environment Interaction Protocol}
\label{subsec:protocol}
As shown in Figure~\ref{fig:overview}, GeoBuildBench is structured as an interactive agent--environment loop.
At each step, the agent observes the problem description and previous environment feedback to
 emit or modify a DSL program without assuming a fixed granularity of updates.
The environment executes the program, renders a  diagram, and returns visual output
together with structured feedback regarding execution errors, missing required objects or violated constraints.

This interaction continues iteratively, allowing the agent to inspect and repair its construction
based on feedback. 
The interaction terminates upon success or when the interaction budget is reached. 
%This protocol enables systematic evaluation of multi-step, visually grounded geometry construction and isolates an agent’s ability to reason about and exploit geometric properties through  auxiliary constructions under strict semantic constraints.

%%%%%%%%%%%%%%%%%%%%%%%%%%%%%%%%%%%%%%%%%%%%%%%%%%%%%%%%%%%%

\section{GeoBuildBench Dataset}
\label{sec:dataset}

% \subsection{Sources and Scope}
% \label{subsec:sources}
The benchmark consists of 489 Chinese plane-geometry problems curated from GeoQA~\cite{chen-etal-2021-geoqa} and additional online textbook repositories. Each instance is represented by its natural-language text paired with a canonicalized set of required objects and a list of verification conditions that encode the semnatic constraints. We do not claim to cover every possible relation type that may appear in plane geometry. Instead, the verification conditions are defined over a practical closed set of relation types that frequently occur in GeoQA-style problems and admit reliable numerical checking (see Appendix~\ref{app:verification-conditions} for the  type list and semantics).
%We focus on a practical set of relation types that admit reliable numerical checking.

% GeoBuildBench is constructed from two sources:
% (i) geometry problems drawn from GeoQA\cite{chen-etal-2021-geoqa}, and
% (ii) additional textbook-style problems collected from online repositories.
% After filtering, annotation, and verification, the final benchmark contains
% 489 Chinese plane-geometry problems.

% Each problem is represented by a natural-language description together with
% (1) a set of required geometric objects and
% (2) a list of verification conditions that encode the semantic constraints
% of the intended geometric construction.
% We do not claim to cover every possible relation type that may appear in plane-geometry text.
% Instead, the verification conditions are defined over a practical closed set of
% relation types that frequently occur in GeoQA-style problems and admit reliable numerical checking
% (see Appendix~\ref{app:verification-conditions} for the full type list and semantics).

\subsection{Curation Pipeline}
\label{subsec:pipeline}
Building a construction-centric benchmark poses challenges beyond those in
standard geometry QA datasets.
Many problems rely on information that is only recoverable from an accompanying
diagram, contain ambiguous references, or specify internally inconsistent constraints.
To ensure that each instance is solvable \emph{from text alone as a geometric construction task}, we apply a three-stage curation pipeline combining LLM-based
processing with human verification.

\paragraph{Stage 1: Text-based constructibility filtering and cleaning.}
We first use GPT-4.1 to assess whether a problem is suitable for geometric construction from text alone. The model is prompted to (i) reject problems whose geometric meaning depends on diagram-only cues (e.g., numbered angles such as ``$\angle 1$''), undefined point
references, or incomplete constraints, and (ii) remove non-constructive content, including answer queries, multiple-choice options, score annotations, and boilerplate phrases such as ``as shown in the figure''.
The output of this stage is a \emph{cleaned text} that retains only geometric setup conditions, e.g., incidences, parallelism, angle relations, and metric constraints.

Notably, a problem may appear well-specified in natural language yet still be inconstructible due to hidden geometric inconsistencies. 
Such cases highlight that geometric constructibility is nontrivial and cannot be determined from surface form alone. We provide concrete examples of both constructible and inconstructible cases in Appendix~\ref{app:examples}.

\paragraph{Stage 2: Task representation extraction.}
For problems that pass Stage 1, we apply a second GPT-4.1 prompt to extract a structured task representation.
This stage converts the cleaned text into
(i) \texttt{required\_objects}, e.g., points, segments/lines, circles, and polygons,
(ii) \texttt{verification\_conditions}, e.g., point-on-segment, angle values, parallelism, length relations, and
(iii) auxiliary metadata such as category labels and an estimated construction difficulty. 
The prompt enforces a fixed schema and a closed set of supported condition types, ensuring consistency across instances.

Importantly, this stage is \emph{not} intended to solve the geometry problem:
it only maps surface geometric descriptions into pre-defined condition templates. 
As a result, the extraction can succeed even when the model's geometric reasoning is weak,
because it primarily requires schema-constrained mapping rather than multi-step deduction. 
Moreover, although the same geometric relation may be expressed in diverse natural-language forms, 
the output is explicitly canonicalized into our finite condition-type set, which later enables unambiguous constraint-based evaluation and hallucination diagnosis.

\paragraph{Stage 3: Human verification.}
Finally, all retained instances undergo human verification. Annotators check that (i) the extracted task representation reflects the cleaned text, (ii)  required objects and conditions are neither missing nor malformed,
and (iii) the constraint set is geometrically realizable, i.e., a valid construction exists in principle. 
Crucially, annotators are \emph{not} asked to re-interpret the full problem or invent new constraints;
their role is primarily to validate that the LLM output conforms to the fixed schema and intended semantics,
and to filter out rare cases of inconsistency (e.g., mutually incompatible length and perimeter constraints).
Problems that are syntactically well-formed but geometrically inconsistent are removed at this stage.

\paragraph{Annotators and consistency.}
Human verification was conducted by two graduate-level annotators with formal
training in mathematics.
Both annotators followed a unified annotation guideline specifying the allowed
object types, verification condition schema, and rejection criteria, and were
instructed to assess geometric realizability strictly under the given constraints,
without introducing unstated assumptions.

To ensure consistency, all instances were reviewed by at least one annotator,
and a subset of cases were independently checked by a second annotator.
Disagreements were rare and were resolved through discussion.
The verification process focuses on checking schema conformity and geometric
realizability, and does not require solving the problems.

\subsection{Task Representation}\label{subsec:task-representation}
Each verified problem is represented as a construction task defined by
\texttt{required\_objects} and \texttt{verification\_conditions}.
Importantly, GeoBuildBench does not assume a single gold diagram or construction
procedure.
Multiple constructions are considered valid as long as they introduce all
required objects and satisfy all verification conditions.
This design explicitly allows auxiliary constructions (e.g., helper points,
lines, or circles), reflecting standard human geometric practice.
An example task instance is shown in Appendix~\ref{app:examples}.

\subsection{Dataset Statistics}
\label{subsec:dataset-statistics}

GeoBuildBench covers a broad range of plane-geometry topics commonly found in
textbook-style problems, with a majority of instances involving circle- and
triangle-based constructions.
The benchmark also includes problems featuring angle relationships, geometric
transformations, metric relations, polygons, and similarity or congruence.

Each problem is annotated with a construction difficulty level ranging from
1 (very easy) to 4 (hard), reflecting the complexity of \emph{constructing}
the geometric configuration rather than solving a downstream numerical or
proof-based question.
Most instances fall into moderate difficulty levels, making the benchmark
well suited for evaluating multi-step construction and error recovery.

Detailed category and difficulty distributions are provided in
Appendix~\ref{app:dataset}.
%%%%%%%%%%%%%%%%%%%%%%%%%%%%%%%%%%%%%%%%%%%%%%%%%%%%%%%%%%%%

\section{Benchmark Evaluation and Analysis}
\label{sec:benchmark-evaluation}

\subsection{Experimental Setup}
\label{subsec:experimental-setup}

We evaluate GeoBuildBench using a set of modern multimodal large language models
capable of reasoning and geometry construction with visual input support.
For each model, we adopt a fixed interaction budget of \textbf{maximum 5 iterations},
where at each step the agent can emit a DSL program to be executed by our geometry
kernel and receive structured feedback.

All experiments use the same execution environment, ensuring fairness across models.
Each problem instance is processed in \emph{vision-enabled mode}, whereby the
agent observes both the textual problem description and a rendering of the
current geometric state.

We test the following models:
\begin{itemize}
  \item \textbf{GPT-5.1} (OpenAI): state-of-the-art reasoning and multimodal LLM \cite{openai_gpt51}.

  \item \textbf{Gemini-3-Flash} (Google) — low-latency reasoning model \cite{google_gemini3flash}.

  % \item \textbf{GPT-5 Mini} (OpenAI): lightweight variant of GPT-5 optimized for efficiency and cost-sensitive deployment.
  
  % \item \textbf{Gemini-2.5-Pro} (Google): high-capacity multimodal reasoning model with stronger performance on complex tasks.
  
  % % \item \textbf{Gemini-2.5-Flash} (Google): low-latency multimodal model designed for fast reasoning and interactive settings.

  % \item \textbf{Claude Sonnet 4.5} (Anthropic): advanced reasoning-oriented large language model with strong instruction-following capabilities.

  \item \textbf{Qwen3-VL-235B-A22B-Instruct} (Alibaba):  advanced multimodal vision-language model \cite{qwen3vl2025}.
  \item \textbf{LLaMA-3.2-90B-Vision-Instruct} (Meta):  open-weight vision reasoning model \cite{meta_llama3_2_vision}.
\end{itemize}

During evaluation, each model is prompted to construct the geometry in an
iterative loop. We do not finetune models; all evaluations use zero-shot or
few-shot prompting strategies consistent across models.

\subsection{Evaluation Metrics}
\label{subsec:evaluation-metrics}

To assess model performance on GeoBuildBench, we define the following metrics:

\paragraph{Success Rate}
The fraction of problems where the agent successfully constructs a diagram
satisfying all required objects and verification conditions within the fixed
interaction budget.

\paragraph{Iteration Steps}
We report the distribution of steps used to reach successful solutions, including
the minimum, average, and maximum number of iterations.

\paragraph{Hallucination Analysis}
We count hallucination events: cases where the model references non-existent
objects, emits syntactically invalid DSL, or returns mismatched program output.
We also analyze the average number of hallucinations per problem and
recovery rates.

\paragraph{Missing Objects}
We track the total number of geometric objects that were required by a task
but not produced by the model’s construction program, broken down by type
(e.g., points, segments, circles).

\paragraph{Failed Conditions}
We measure the total number of verification conditions that remain unsatisfied
after program execution. We classify failures by condition type
(e.g., \texttt{angle\_value}, \texttt{parallel}, etc.).

These metrics together provide a comprehensive view of a model’s ability to
ground textual geometry problems into executable constructions.

\subsection{Overall Results}
\label{sec:overall-results}

Table~\ref{tab:main-results} reports the overall performance of four
state-of-the-art vision-language models on GeoBuildBench.
A construction is considered successful only if it executes without error,
covers all required geometric objects, and satisfies all verification conditions.

\paragraph{Additional analyses.} To further probe language effects and the role of problem difficulty, we perform two supplementary experiments: (i) evaluation on an English translation of GeoBuildBench and (ii) an analysis of success counts by difficulty level.  The corresponding results are reported in Appendix Tables~\ref{tab:english-results} and~\ref{tab:difficulty-analysis}, respectively.

\begin{table*}[t]
\caption{
Overall performance on GeoBuildBench.
All models are evaluated with a maximum of 5 interaction steps.
Lower is better ($\downarrow$) for steps, hallucinations, missing objects, and
failed constraints.
}
\label{tab:main-results}
\centering
\small
\begin{tabular}{lccccc}
\toprule
Model &
Success Rate (\%) &
Avg. Steps $\downarrow$ &
\makecell{Hallucinations \\ / Prob. $\downarrow$} &
\makecell{Missing \\ Objects $\downarrow$} &
\makecell{Failed \\ Constraints $\downarrow$} \\
\midrule
GPT-5.1
& \textbf{78.9}
& 1.87
& 0.40
& 939
& 1119 \\

Gemini-3-Flash
& 75.3
& \textbf{1.55}
& \textbf{0.34}
& \textbf{329}
& \textbf{932} \\

Qwen3-VL-235B
& 42.2
& 2.30
& 2.30
& 2042
& 1817 \\

LLaMA-3.2-90B-Vision
& 21.3
& 2.23
& 2.38
& 1823
& 1584 \\
\bottomrule
\end{tabular}
\end{table*}

\begin{table*}[t]
\caption{
Structural hallucination frequency and recovery behavior on GeoBuildBench.
Lower is better ($\downarrow$) for all metrics.
}
\label{tab:hallucination-recovery}
\centering
\small
\begin{tabular}{lccc}
\toprule
Model &
\makecell{Hallucinations \\ per Problem $\downarrow$} &
\makecell{Recovery Steps \\ per Hallucination $\downarrow$} &
\makecell{Avg. Success \\ Steps $\downarrow$} \\
\midrule
GPT-5.1
& 0.40
& 1.29
& 1.87 \\

Gemini-3-Flash
& \textbf{0.34}
& \textbf{1.17}
& \textbf{1.55} \\

Qwen3-VL-235B
& 2.30
& 1.74
& 2.30 \\

LLaMA-3.2-90B-Vision
& 2.38
& 1.79
& 2.23 \\
\bottomrule
\end{tabular}
\end{table*}

We observe substantial variation across models.
GPT-5.1 and Gemini-3-Flash achieve the highest success rates
(78.9\% and 75.3\%, respectively),
indicating that current frontier MLLMs can often construct valid geometric
configurations under explicit semantic constraints.
However, these two models exhibit different error profiles.
Gemini-3-Flash requires fewer steps on average
(1.55 vs.\ 1.87) and produces the lowest number of hallucinations per problem
(0.34),
suggesting more stable geometric construction behavior,
while GPT-5.1 achieves a slightly higher overall success rate.

In contrast, Qwen3-VL-235B and LLaMA-3.2-90B-Vision perform significantly worse,
with success rates of 42.2\% and 21.3\%, respectively.
Both models exhibit substantially higher hallucination rates
(above 2.3 per problem on average),
along with large numbers of missing required objects and failed geometric
constraints.
These failures indicate difficulty in maintaining consistent geometric state
and respecting relational constraints over multi-step construction.

Across all models, we find that unsuccessful cases are dominated by structural
errors rather than execution failures.
In particular, missing geometric objects and violations of angle, incidence,
and metric constraints account for the majority of failures,
highlighting that GeoBuildBench primarily probes semantic grounding and
structural reasoning, rather than surface-level syntax or program execution.

\subsection{Structural Hallucinations}
\label{sec:structural-hallucinations}

We first analyze \emph{structural hallucinations}, which capture failures in
maintaining a coherent geometric state during construction.
We define a structural hallucination as an error that introduces undefined
geometric references, syntactically invalid DSL programs, or mismatches between
declared and produced objects.

Table~\ref{tab:hallucination-recovery} shows that hallucination frequency varies
substantially across models.
Gemini-3-Flash and GPT-5.1 exhibit low hallucination rates
(0.34 and 0.40 per problem, respectively),
while Qwen3-VL-235B and LLaMA-3.2-90B-Vision produce more than
two hallucinations per problem on average.

Across all models, the most frequent hallucination type is
\emph{undefined reference}, where a DSL command refers to a geometric object
that has not been instantiated.
This trend holds consistently across models
(see Appendix~\ref{app:analysis-hallucination}),
with weaker models exhibiting an order-of-magnitude increase in such errors.
This indicates that most failures arise from difficulties in tracking and
reusing previously constructed geometric objects, rather than from superficial
syntax issues.
The prevalence of such errors highlights the challenge of maintaining a
consistent internal object graph over multi-step geometric construction.

A detailed quantitative breakdown of hallucination types, missing objects,
and failed verification conditions is provided in
Appendix~\ref{app:analysis-failed-conditions}.

\begin{table*}[t]
\centering
\small
\caption{Vision ablation on open-weight models.}
\label{tab:vision_ablation}
\begin{tabular}{llccccc}
\toprule
Model & Setting & Success Rate (\%) & Avg. Steps $\downarrow$ & \makecell{Missing \\ Objects} $\downarrow$ & \makecell {Failed \\ Constraints} $\downarrow$ & Halluc. $\downarrow$ \\
\midrule
\multirow{2}{*}{Qwen3-VL-235B}
& with vision & 42.2 & 2.30 & 2042 & 1817 & 928 \\
& no vision   & 39.1 & 2.08 & 131  & 100  & 85  \\
\midrule
\multirow{2}{*}{LLaMA-3.2-90B-Vision}
& with vision & 21.3 & 2.23 & 1823 & 1584 & 819 \\
& no vision   & 23.1 & 2.19 & 2408 & 2165 & 1108 \\
\bottomrule
\end{tabular}
\end{table*}

\subsection{Feedback Sensitivity and Error Recovery}
\label{sec:feedback-recovery}

GeoBuildBench explicitly exposes agents to visual renderings and structured
constraint feedback after each construction attempt.
This enables analysis not only of how often models make structural errors, but
also of how effectively they recover from them.

We quantify recovery behavior using the average number of steps required to
resolve a hallucination once it occurs.
As shown in Table~\ref{tab:hallucination-recovery}, frontier models recover
quickly from structural errors.
Gemini-3-Flash achieves the fastest recovery
(1.17 steps per hallucination),
followed by GPT-5.1 (1.29 steps),
indicating that these models often correct errors immediately after receiving
feedback.

In contrast, Qwen3-VL-235B and LLaMA-3.2-90B-Vision exhibit substantially weaker
recovery behavior, requiring 1.74 and 1.79 steps per hallucination, respectively.
Despite being allowed the same interaction budget, hallucinations in these
models frequently persist across iterations rather than being resolved.

Importantly, hallucination frequency and recovery speed capture complementary
aspects of agent behavior.
Some models exhibit frequent structural errors but are nevertheless able to
interpret feedback and repair them efficiently.
Others fail both to maintain a stable geometric state and to exploit feedback
for correction.
These results demonstrate that iterative interaction alone is insufficient:
effective geometry construction requires sensitivity to explicit structural
feedback and the ability to update internal representations accordingly.

\subsection{Auxiliary Constructions and Geometric Flexibility}
\label{sec:auxiliary-constructions}

A distinctive feature of GeoBuildBench is that it permits auxiliary geometric
objects and constructions that are not explicitly mentioned in the problem
statement.
This design mirrors standard human geometric practice, where additional points,
circles, or perpendiculars are often introduced to satisfy metric or angular
constraints.

We observe that successful models frequently exploit this flexibility.
Both GPT-5.1 and Gemini-3-Flash regularly introduce helper constructions, such as
circles to enforce length constraints or perpendicular lines to realize right
angles.
These auxiliary objects are retained in the construction state but ignored by
the verification module as long as all required objects and constraints are
satisfied.

In contrast, weaker models rarely make effective use of auxiliary constructions.
Instead, they tend to omit required relations or repeatedly attempt to satisfy
constraints without constructing the necessary geometric support.
This behavior suggests a more superficial treatment of geometric relations and
an inability to flexibly leverage geometric properties.

By allowing auxiliary constructions, GeoBuildBench evaluates not whether a model
reproduces a canonical diagram, but whether it can \emph{use geometry as a
toolkit} to realize constraints.
This capability is largely invisible to benchmarks based on visual similarity or
single-shot diagram generation.

\subsection{Ablation Study}

We conduct a controlled vision ablation on the two open-weight models. For each model, we toggle image inputs (\textit{with vision} vs.\ \textit{no vision}) while keeping the agent--environment loop, interaction budget, and prompting template fixed. Table~\ref{tab:vision_ablation} summarizes the ablation outcomes. Overall, enabling vision produces \emph{small} changes in headline success rates, but \emph{large} shifts in failure-mode profiles (missing objects, failed conditions, and structural hallucinations).

For Qwen3-VL, enabling vision yields a modest success increase (42.2\% vs.\ 39.1\%), but substantially increases all verifier-diagnosed error signals: missing required objects (2042 vs.\ 131), unsatisfied conditions (1817 vs.\ 100), and structural hallucinations (928 vs.\ 85). This suggests that vision can help Qwen identify viable construction ideas more often, but it simultaneously destabilizes symbol binding and state tracking (e.g., referencing objects not yet created or not present in the current
construction state).

For LLaMA-3.2-Vision, enabling vision does not improve success (21.3\% vs.\ 23.1\%), but it consistently reduces structural errors: missing objects (1823 vs.\ 2408), failed conditions (1584 vs.\ 2165), and hallucinations (819 vs.\ 1108). This pattern indicates that vision primarily improves output cleanliness and object referencing correctness, while the dominant remaining bottleneck is geometric planning and constraint satisfaction.

\subsection{Discussion}\label{sec:discussion}

Taken together, our results show that performance on GeoBuildBench is governed
primarily by a model’s ability to maintain a coherent geometric state, recover
from structural errors using explicit feedback, and flexibly employ auxiliary
constructions.
High success rates correlate with low hallucination frequency and fast recovery,
rather than with raw model size or general vision-language capacity.

A key implication is that many failure modes revealed by GeoBuildBench would likely be missed by evaluation protocols that rely only on visual plausibility or  answer correctness. A diagram can look reasonable to a human observer while  violating essential geometric relations, and an agent may reach the correct answer despite an internally inconsistent or incomplete construction. In contrast, GeoBuildBench makes these errors explicit by verifying object existence and constraint satisfaction.

Overall, these findings suggest that current MLLMs—despite strong performance on conventional geometry tasks—still struggle with robust, executable geometry construction. GeoBuildBench offers a controlled testbed for diagnosing such limitations and for supporting future work on structurally grounded, agentic reasoning for geometry.

Additional breakdowns of hallucination categories, missing-object cases, and violated verification conditions are reported in the Appendix.

%%%%%%%%%%%%%%%%%%%%%%%%%%%%%%%%%%%%%%%%%%%%%%%%%%%%%%%%%%%%

\section*{Limitations}
\label{sec:limitations}

GeoBuildBench is designed to benchmark grounded, executable geometry construction, but it does not fully cover the space of geometric language understanding or geometric reasoning. The benchmark scope is intentionally constrained. Tasks focus on \emph{Chinese} plane-geometry problems and a practical, closed set of required-object types and verification condition templates. As a result, performance on GeoBuildBench may not transfer directly to (i) other languages (e.g., English problem statements), (ii) broader Euclidean topics that rely on relations outside our condition-type set, or (iii) settings that require richer mathematical context (e.g., analytic geometry, 3D geometry, or non-Euclidean variants). Relatedly, the action space is defined by our Geometry Construction DSL. While it is expressive for common constructions, it remains a DSL abstraction rather than a full-featured dynamic geometry system, and agent behaviors may differ under alternative tooling or interaction primitives. Moreover, the reported model results are contingent on a particular interactive protocol (vision-enabled feedback, fixed iteration budget, and a specific prompting and harness design). Different step budgets, feedback granularity, or prompting strategies may change absolute success rates and error profiles. We therefore recommend interpreting the reported numbers primarily as evidence about failure modes and relative robustness under our controlled setting, rather than as definitive rankings across all geometry-construction regimes.

% \section*{Acknowledgments}

\bibliography{custom}

%%%%%%%%%%%%%%%%%%%%%%%%%%%%%%%%%%%%%%%%%%%%%%%%%%%%%%%%%%%%
\newpage
\appendix
\clearpage

\section{GeoDSL Quick Reference}
\label{app:geodsl}

\paragraph{Syntax.}
Each command follows:
\[
\texttt{command : inputs -> outputs}
\]
Commands are executed sequentially. Outputs are named objects that can be
referenced by later commands.

\paragraph{Numeric literals and expressions.}
GeoDSL supports inline numeric literals (integers and floats) and
arithmetic/trigonometric expressions for coordinates, radii, and angles.
Supported operators/functions include \texttt{+,-,*,/,( )} and
\texttt{sin}, \texttt{cos}, \texttt{tan}.
Angles can be specified in degrees (e.g., \texttt{90°}, \texttt{45deg}) or radians
(e.g., \texttt{1.5708rad}, \texttt{1.5708r}); unitless angles default to degrees.

\paragraph{Design note.}
GeoDSL is a \emph{construction} language: it provides commands to build geometric
objects rather than to assert constraints.
Auxiliary constructions (additional points, lines, or circles) are allowed.

%------------------------------------------------
\subsection{Primitive Object Constructors}
\label{app:geodsl-primitives}

Table~\ref{tab:geodsl-primitives} summarizes the primitive commands for creating
geometric objects in GeoDSL, including points, lines, segments, rays, circles,
and angle objects.
All primitives deterministically create concrete geometric entities that can be
referenced in subsequent commands.

\begin{table*}[t]
\caption{Primitive object constructors in GeoDSL.}
\label{tab:geodsl-primitives}
\centering
\small
\renewcommand{\arraystretch}{1.25}
\begin{tabular}{p{2.3cm}p{2.8cm}p{2.5cm}p{5.2cm}}
\toprule
\textbf{Category} & \textbf{Command} & \textbf{Inputs} & \textbf{Outputs / Description} \\
\midrule
Point &
\texttt{point} &
\texttt{--}, \texttt{x y}, \texttt{line}, \texttt{circle} &
Creates a point. Coordinates create a fixed point; line/circle inputs sample a
point on the object. \\

Line &
\texttt{line} &
\texttt{A B} &
Line through points $A$ and $B$. \\

Line &
\texttt{line\_bisector} &
\texttt{A B} &
Perpendicular bisector of segment $AB$. \\

Segment &
\texttt{segment} &
\texttt{A B} &
Segment with endpoints $A$ and $B$. \\

Ray &
\texttt{ray} &
\texttt{A B} &
Ray starting at $A$ and passing through $B$. \\

Circle &
\texttt{circle} &
\texttt{O A}, \texttt{O r} &
Circle centered at $O$, defined by a point or radius. \\

Angle &
\texttt{angle} &
\texttt{A B C} &
Angle at vertex $B$ formed by rays $BA$ and $BC$. \\

Constant &
\texttt{const int}, \texttt{const Measure} &
\texttt{v} &
Creates numeric or length constants. \\
\bottomrule
\end{tabular}
\end{table*}

%------------------------------------------------
\subsection{Construction Operators}
\label{app:geodsl-operators}

Table~\ref{tab:geodsl-ops} lists operators for common geometric constructions,
including intersections, parallel and perpendicular lines, midpoints, and
rotations.

\begin{table*}[t]
\caption{Construction operators in GeoDSL.}
\label{tab:geodsl-ops}
\centering
\small
\renewcommand{\arraystretch}{1.25}
\begin{tabular}{p{3.0cm}p{2.5cm}p{2.5cm}p{5.8cm}}
\toprule
\textbf{Command} & \textbf{Inputs} & \textbf{Outputs} & \textbf{Description} \\
\midrule
\texttt{intersect} &
\texttt{obj1 obj2} &
\texttt{P} or \texttt{P Q} &
Intersection of geometric objects (e.g., line--line, line--circle). \\

\texttt{parallel\_line} &
\texttt{P ref} &
\texttt{l} &
Line through $P$ parallel to a reference object. \\

\texttt{orthogonal\_line} &
\texttt{P line} &
\texttt{l} &
Line through $P$ perpendicular to a given line. \\

\texttt{midpoint} &
\texttt{A B} &
\texttt{M} &
Midpoint of segment $AB$. \\

\texttt{rotate} &
\texttt{P $\theta$ Center} &
\texttt{Q} &
Rotates point $P$ by angle $\theta$ around \texttt{Center}. \\
\bottomrule
\end{tabular}
\end{table*}

%------------------------------------------------
\subsection{Measurement and Arithmetic Utilities}
\label{app:geodsl-math}

Table~\ref{tab:geodsl-math} lists commands for distance measurement and basic
arithmetic operations.
Most arithmetic can alternatively be expressed via inline expressions.

\begin{table*}[t]
\caption{Measurement and arithmetic utilities in GeoDSL.}
\label{tab:geodsl-math}
\centering
\small
\renewcommand{\arraystretch}{1.25}
\begin{tabular}{p{2.5cm}p{2.3cm}p{2.3cm}p{5.1cm}}
\toprule
\textbf{Command} & \textbf{Inputs} & \textbf{Outputs} & \textbf{Description} \\
\midrule
\texttt{distance} & \texttt{A B} & \texttt{d} & Distance between points $A$ and $B$. \\
\texttt{sum} & \texttt{a b} & \texttt{c} & Addition. \\
\texttt{minus} & \texttt{a b} & \texttt{c} & Subtraction. \\
\texttt{product} & \texttt{a b} & \texttt{c} & Multiplication. \\
\texttt{ratio} & \texttt{a b} & \texttt{c} & Division. \\
\bottomrule
\end{tabular}
\end{table*}

%%%%%%%%%%%%%%%%%%%%%%%%%%%%%%%%%%%%%%%%%%%%%%%%%%%%%%%%%%%%%%%%%%%%%%%
\section{Verification Semantics and Failure Modes}
\label{app:verification}

This appendix formally specifies the geometric verification semantics used in
GeoBuildBench.
It defines the scope of supported verification conditions, the object-level
coverage checks applied to each construction, and the interpretation of
verification failures as distinct forms of structural hallucination.

\subsection{Verification Scope and Object Coverage}
\label{app:verification-scope}

Each task instance specifies a set of \emph{required geometric objects} and a
set of \emph{verification conditions}.
Verification proceeds only after executing the generated GeoDSL program and
instantiating a concrete geometric state.

GeoBuildBench does not aim to cover all possible geometric relations that may
appear in natural-language problems.
Instead, verification is defined over a finite and explicitly enumerated set of
condition types that (i) frequently occur in textbook-style plane geometry
problems and (ii) admit reliable numerical evaluation.
This closed design ensures deterministic checking and consistent interpretation
across instances.

Before evaluating relational constraints, the verifier first checks
\emph{object coverage}, ensuring that all required geometric entities are
explicitly present in the final construction.
The object types considered include:
(i) points,
(ii) segments,
(iii) lines,
(iv) circles, and
(v) polygonal objects when specified by the task.

Missing objects are treated as verification failures regardless of whether
auxiliary constructions are permitted.
This separation allows the benchmark to distinguish failures caused by omitted
entities from those arising from incorrect geometric relations.

\subsection{Supported Verification Condition Types}
\label{app:verification-conditions}

GeoBuildBench evaluates a finite and explicitly defined set of verification
condition types.
Although natural-language geometry descriptions may express the same relation
in diverse surface forms, all conditions are normalized into this fixed set
during dataset curation (see Section~\ref{subsec:pipeline}).
As a result, downstream verification and hallucination diagnosis operate on
canonicalized geometric semantics rather than on raw linguistic expressions.

The supported condition types cover a broad range of geometric semantics and can
be grouped into the following categories.

\paragraph{Incidence and topological relations.}
These conditions verify whether geometric objects satisfy required incidence or
ordering relations, including point-on-line, point-on-segment, point-on-circle,
collinearity, concurrency, concyclicity, order-on-line, and related variants
(e.g., extensions or arc-based relations).

\paragraph{Metric and angular constraints.}
Metric conditions enforce equality, sum, or ratio relations among distances,
segments, or angles.
These include distance equality, segment equality, angle value, angle equality,
angle sum, angle ratio, and segment ratio.
Specialized constraints such as isosceles triangle, right triangle, and
perimeter constraints are treated as derived metric relations.

\paragraph{Directional and structural relations.}
Parallelism, perpendicularity, perpendicular bisectors, and diameter-based
conditions are verified as structural relations that constrain the global
configuration of objects.
Violations typically indicate geometrically infeasible constructions rather than
local numerical error.

\paragraph{Curvature and tangency relations.}
Tangency-related conditions, including tangent, tangent at point,
and line-is-tangent, verify first-order contact between linear and circular
objects under numerical tolerance.

\paragraph{Polygonal and higher-level properties.}
When polygonal objects are specified, the verifier checks polygon type, regular
polygon constraints, square constraints, and associated polygon properties.
These conditions are evaluated through their induced metric and angular
relations rather than through symbolic assertions.

\subsection{Numerical Tolerances and Equivalence Handling}
\label{app:verification-tolerance}

All metric and angular conditions are evaluated under fixed numerical tolerances
to account for floating-point imprecision.
For angular constraints, both the measured angle and its reflex
($360^\circ - \theta$) are treated as equivalent to avoid spurious failures due
to orientation ambiguity.
Distance-based conditions are normalized to be invariant to global scaling
applied during rendering.

\subsection{Interpretation of Verification Failures}
\label{app:verification-failure-modes}

Verification failures are interpreted as evidence of distinct structural failure
modes in the agent--environment loop.
Importantly, because all conditions are expressed in a canonicalized form,
failure diagnosis is independent of the original linguistic realization of the
constraint.

\paragraph{Non-existent object references.}
Conditions that reference undefined geometric entities indicate hallucinated
symbols that were never constructed in the executable geometry.

\paragraph{Missing required constructions.}
Failures of object coverage or incidence relations reflect omissions of required
points, segments, lines, or circles, even when auxiliary constructions are
allowed.

\paragraph{Geometrically infeasible relations.}
Violations of parallelism, perpendicularity, tangency, or angle constraints
typically arise from internally inconsistent or infeasible configurations that
cannot be satisfied simultaneously.

\paragraph{Semantic mismatch under tolerance.}
Near-satisfied metric or angular constraints that fall outside numerical
tolerance indicate partial semantic understanding rather than syntactic or
execution failure.

By explicitly separating object coverage, canonicalized relational verification,
and numerical evaluation, GeoBuildBench enables semantics-aware diagnostics that
go beyond textual correctness or visual similarity and directly probe structured
geometric reasoning.
%%%%%%%%%%%%%%%%%%%%%%%%%%%%%%%%%%%%%%%%%%%%%%%%%%%%%%%%%%%%%%%%%%%%%%%

% \clearpage
\lstdefinestyle{promptstyle}{
  basicstyle=\ttfamily\footnotesize,
  columns=fullflexible,
  keepspaces=true,
  breaklines=true,
  breakatwhitespace=false,
  frame=single,
  framerule=0.4pt,
  rulecolor=\color{black},
  xleftmargin=0pt,
  xrightmargin=0pt,
  aboveskip=0.6em,
  belowskip=0.6em,
  literate=
    {°}{{\textdegree}}1
    {✅}{{[OK]}}1
    {❌}{{[NO]}}1
    {θ}{{theta}}1
    {≈}{{$\sim$}}1
    {∠}{{$\angle$}}1
}

% ------------------------------------------------------------
\section{Full Agent System Prompt}
\label{app:full-agent-prompt}

\lstset{style=promptstyle}

\begin{lstlisting}[
  caption={Full system prompt used for all agents in GeoBuildBench.},
  label={lst:full-agent-prompt}
]

You are a geometry problem-solving agent that uses a Domain-Specific Language (DSL) to construct geometric figures.

Your task is to read Chinese geometry problems and generate DSL code that creates the geometric construction described in the problem.

## DSL Syntax Overview

Format: command : inputs -> outputs

### Basic Commands
# Points
point : 100 150 -> P

# Mathematical Expressions (NO SPACES!)
point : 50+30 100-20 -> Q
point : cos(30°) sin(30°) -> R
point : 100*cos(45°) 100*sin(45°) -> S
point : (100+100*cos(115°)) (0+100*sin(115°)) -> P

# Lines & Segments
line : A B -> line_AB
segment : A B -> seg_AB
ray : A B -> ray_AB

# Circles
circle : O A -> circle_O
circle : O 50 -> circle_O
circle : O 100*sin(60°) -> c

# Angles & Rotation
angle : A B C -> angle_ABC
rotate : P 60 Center -> P_rot
rotate : P 60° Center -> P_rot

# Constructions
midpoint : A B -> M
orthogonal_line : P line -> perp
parallel_line : P line -> para
intersect : line1 line2 -> P
intersect : line circle_O -> P1 P2
line_bisector : A B -> bisector
angular_bisector : A B C -> bis

# Triangle Centers
incenter : A B C -> I
circumcenter : A B C -> O
incircle : A B C -> circle_I
circumcircle : A B C -> circle_O

## CRITICAL RULES
1. Define before use
2. One definition per label
3. NO polygon command
4. Construct, don't assert
5. Use expressions for precision

\end{lstlisting}

% \newpage
\section{Task Parsing and Annotation Prompts}
\label{app:task-prompts}

This appendix reports the exact prompts used to filter, clean, and annotate
Chinese geometry problems for inclusion in GeoBuildBench.
These prompts define the construction suitability criteria and the formal
extraction of geometric objects and verification conditions.
\newpage
\subsection{Construction Suitability Filtering Prompt}
\label{app:prompt-filter}

\lstset{style=promptstyle}
\begin{lstlisting}
Analyze this geometry problem for GEOMETRIC CONSTRUCTION suitability.

Problem: {problem_text}

Your task:
1. Determine whether this problem is suitable for geometric figure construction
2. Remove non-construction content (questions, scores, diagram references, etc.)
3. Keep ONLY the geometric setup conditions

REJECTION CRITERIA (return is_valid: false if ANY apply):

1. Undefined Points:
   - "angle E = 40 degrees" (point E is not geometrically defined)
   - "angle BDC = 30 degrees" (point D is not defined)
   - Valid: "point D lies on segment AB, angle BDC = 30 degrees"

2. Ambiguous Angles:
   - "angle D = 26 degrees"
   - Valid: "angle ABC = 50 degrees"

3. Diagram-Dependent Angle Labels:
   - "angle 1 = 30 degrees, angle 2 = 45 degrees"

4. Incomplete Constraints:
   - "AB is parallel to CD, angle E = 40 degrees" (angle location undefined)

5. Pure Calculation Problems:
   - Problems that only ask for numerical values without defining a constructible figure

CLEANING RULES:
- Remove score indicators
- Remove diagram references (e.g., references to figures)
- Remove questions or proof requests
- Remove multiple-choice answers
- Remove any text appearing after result or query markers
  (e.g., result clauses, questions, or proof statements)

KEEP:
- Shape definitions (e.g., triangles, quadrilaterals)
- Explicit point positions
- Measurements (lengths, angles)
- Geometric relations (parallel, perpendicular, etc.)

Return JSON:
{
  "is_valid": true/false,
  "cleaned_text": "geometry construction conditions only",
  "rejection_reason": ""
}

Only return the JSON.
\end{lstlisting}

\newpage
\subsection{Formal Construction Annotation Prompt}
\label{app:prompt-annotation}

\lstset{style=promptstyle}
\begin{lstlisting}
Parse this geometry problem and extract construction requirements.

Problem: {problem_text}

TASK 1: Extract geometric objects and conditions
TASK 2: Classify into ONE category
TASK 3: Rate construction difficulty (1-5)

Return JSON:
{
  "required_objects": {
    "points": [],
    "segments": [],
    "lines": [],
    "circles": [],
    "polygons": []
  },
  "verification_conditions": [],
  "category": "...",
  "difficulty": 3
}

SUPPORTED CONDITION TYPES:
- parallel, perpendicular, collinear, concurrent
- angle_value, angle_equality, angle_sum, angle_ratio
- segment_equality, segment_ratio, perimeter
- point_on_segment, midpoint_of, order_on_line
- point_on_circle, tangent_line, diameter
- triangle_valid, isosceles_triangle, right_triangle
- polygon_property, polygon_type, square, regular_polygon

CRITICAL RULES:
- All points must be explicitly defined
- Use only supported condition types
- difficulty reflects construction complexity, not solving difficulty

Only return JSON.
\end{lstlisting}

\section{Dataset Statistics and Annotation Details}
\label{app:dataset}

This appendix provides detailed statistics and annotation information for
GeoBuildBench, complementing the high-level description in the main text.
We report the distribution of problem categories and construction difficulty
levels, and clarify the principles used during difficulty annotation.

\subsection{Category Distribution}
\label{app:dataset-category}

GeoBuildBench spans a diverse range of plane-geometry topics commonly appearing
in middle- and high-school textbooks.
Each problem is assigned to a single high-level category based on its primary
geometric structure and constraints.

Table~\ref{tab:category-dist} reports the distribution over categories.
Circle- and triangle-based constructions constitute the majority of the dataset,
while a substantial portion involves angle relationships, geometric
transformations, and metric constraints.
Polygonal constructions and similarity or congruence relations are also included,
ensuring coverage beyond simple primitive configurations.

\begin{table}[H]
\centering
\caption{Category distribution of GeoBuildBench (489 problems).}
\label{tab:category-dist}
\small
\begin{tabular}{lrr}
\toprule
Category & Count & Ratio (\%) \\
\midrule
Circle & 154 & 31.5 \\
Triangle & 143 & 29.2 \\
Angle Relationships & 80 & 16.4 \\
Geometric Transformations & 41 & 8.4 \\
Polygon & 32 & 6.5 \\
Metric Relations & 25 & 5.1 \\
Similarity \& Congruence & 9 & 1.8 \\
Basic Constructions & 5 & 1.0 \\
\bottomrule
\end{tabular}
\end{table}

\subsection{Construction Difficulty Annotation}
\label{app:dataset-difficulty}

Each problem in GeoBuildBench is annotated with a discrete construction
difficulty level ranging from 1 (Very Easy) to 4 (Hard).
Importantly, this difficulty score reflects the complexity of \emph{constructing}
the geometric configuration itself, rather than the difficulty of solving any
associated numerical or proof-based question.

Difficulty levels are assigned during dataset curation based on the following
factors:
(i) the number of required geometric objects,
(ii) the degree of dependency between constructions (e.g., intersection-defined
points or chained auxiliary constructions),
and (iii) the precision and interaction of metric and angular constraints.
Problems that require multiple auxiliary constructions or tight constraint
coordination are rated as higher difficulty.

\begin{table}[H]
\centering
\caption{Construction difficulty distribution in GeoBuildBench.}
\label{tab:difficulty-dist}
\small
\begin{tabular}{lrr}
\toprule
Difficulty Level & Count & Ratio (\%) \\
\midrule
Level 1 (Very Easy) & 54  & 11.0 \\
Level 2 (Easy)      & 238 & 48.7 \\
Level 3 (Medium)    & 162 & 33.1 \\
Level 4 (Hard)      & 35  & 7.2 \\
\bottomrule
\end{tabular}
\end{table}

\subsection{Interpretation and Use in Evaluation}
\label{app:dataset-interpretation}

The distribution shows that GeoBuildBench is dominated by Easy and Medium
instances, reflecting the structure of typical educational geometry problems.
At the same time, the inclusion of a non-trivial fraction of Hard instances
enables stress-testing of multi-step construction, auxiliary reasoning, and
error recovery under feedback.

During evaluation, difficulty labels are not exposed to the models.
Instead, they are used only for analysis, allowing us to examine how construction
complexity correlates with success rate, hallucination frequency, and feedback
utilization.
%%%%%%%%%%%%%%%%%%%%%%%%%%%%%%%%%%%%%%%%%%%%%%%%%%%%%%%%%%%%%%%%%%%%%%%

% \clearpage
\section{Illustrative Construction Examples}
\label{app:examples}

This appendix presents two representative examples to illustrate the
nontrivial nature of geometric construction from natural-language descriptions.
The first example demonstrates a problem that is constructible but requires
non-obvious auxiliary constructions.
The second example highlights a problem that appears well-specified in text
yet is geometrically inconsistent and therefore inconstructible.
Together, these examples clarify both the capabilities and the limitations
addressed by GeoBuildBench.

%------------------------------------------------

\subsection{Example A: Constructible (Tangent--Diameter Configuration)}
\label{app:examples-constructible}

\paragraph{Problem (translated).}
\emph{Let $PA$ and $PB$ be tangents to circle $O$ at points $A$ and $B$.
Let $AC$ be a diameter. Given $\angle APB=40^\circ$.}

\paragraph{Why this example is nontrivial in our setting.}
GeoDSL is a \emph{construction} language (construct, don't assert): it cannot
directly assert ``$PA$ is tangent'' or ``$\angle APB=40^\circ$''.
Instead, the agent must explicitly construct a concrete configuration that
\emph{realizes} tangency (via perpendicularity to radii) and matches the target
angle, potentially introducing auxiliary objects.

\paragraph{A normalized executable instance (one of many valid constructions).}
This problem does not uniquely determine a single diagram up to coordinates.
Therefore, the agent may instantiate any concrete configuration that satisfies
the constraints.
A convenient choice uses the tangent--tangent angle identity:
\[
\angle APB = 180^\circ - \angle AOB.
\]
Thus, enforcing $\angle APB=40^\circ$ can be achieved by setting
$\angle AOB=140^\circ$ in a \emph{normalized coordinate system}.
Importantly, this $140^\circ$ is \emph{not} a literal number read from the text;
it is an intermediate value introduced to instantiate one valid configuration.

\subsubsection*{Executable GeoDSL construction}
\begin{lstlisting}[style=promptstyle,caption={A valid GeoDSL construction for Example A (normalized instance).},label={lst:ex-constructible-dsl}]
point : 0 0 -> O
point : 100 0 -> A
point : 100*cos(140°) 100*sin(140°) -> B

circle : O A -> circle_O
line : O A -> line_OA
line : O B -> line_OB

orthogonal_line : A line_OA -> tangent_A
orthogonal_line : B line_OB -> tangent_B
intersect : tangent_A tangent_B -> P

rotate : A 180° O -> C
segment : P A -> PA
segment : P B -> PB
segment : A C -> AC
\end{lstlisting}

\begin{figure}[H]
  \centering
  \includegraphics[width=0.50\linewidth]{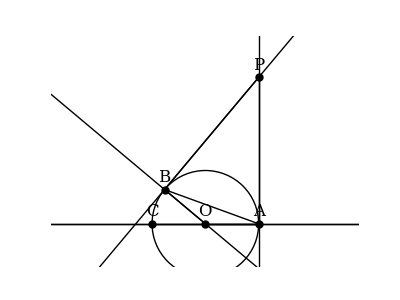}
  \caption{Example A (constructible): a rendered diagram from the GeoDSL program in Listing~\ref{lst:ex-constructible-dsl}.
  The coordinates instantiate one normalized configuration; correctness is judged by verified geometric relations
  (tangency via perpendicularity to radii, and $\angle APB=40^\circ$), not by matching a unique canonical layout.}
  \label{fig:ex-constructible}
\end{figure}

\paragraph{Rendered diagram.}
Figure~\ref{fig:ex-constructible} shows the diagram rendered by our execution engine.

%------------------------------------------------
\subsection{Example B: Well-Formed but Inconstructible}
\label{app:example-inconstructible}

\paragraph{Problem.}
\emph{
In triangle $\triangle ABC$, $BD$ is a median.
Given $AB = 5$, $BC = 3$, and the perimeter of triangle $\triangle ABD$ equals $11$,
determine the perimeter of triangle $\triangle BCD$.
}

\paragraph{Claim.}
There exists no triangle $\triangle ABC$ satisfying all of the following
conditions simultaneously:
\begin{itemize}
    \item $BD$ is a median of $\triangle ABC$ (so $D$ is the midpoint of $AC$),
    \item $AB = 5$,
    \item $BC = 3$,
    \item the perimeter of $\triangle ABD$ is $11$.
\end{itemize}

\paragraph{Proof.}
Since $BD$ is a median, point $D$ is the midpoint of $AC$, and hence
\[
AD = DC = \frac{AC}{2}.
\]
Let
\[
AD = x \quad \text{and} \quad BD = 6 - x,
\]
which follows from the perimeter condition of $\triangle ABD$:
\begin{align*}
& AB + BD + AD = 11 \\
&\Rightarrow 5 + BD + AD = 11 \\
&\Rightarrow BD + AD = 6
\end{align*}

Applying Apollonius' theorem to triangle $\triangle ABC$ with median $BD$, we obtain
\[
AB^2 + BC^2
=
2\left(BD^2 + \left(\frac{AC}{2}\right)^2\right).
\]
Substituting the known values,
\[
25 + 9
=
2\left((6 - x)^2 + x^2\right).
\]
Simplifying yields
\[
2x^2 - 12x + 19 = 0.
\]

The discriminant of this quadratic equation is
\[
\Delta = (-12)^2 - 4 \cdot 2 \cdot 19 = 144 - 152 = -8 < 0.
\]
Since the discriminant is negative, the equation has no real solutions.
Therefore, no real values of $AD$ and $BD$ satisfy all given constraints.

\paragraph{Conclusion.}
No triangle $\triangle ABC$ exists that satisfies the stated conditions.
Although the problem appears coherent and complete in natural language,
its geometric constraints are internally inconsistent.
Such problems are explicitly excluded during the dataset curation process
described in Section~\ref{subsec:pipeline}.

%%%%%%%%%%%%%%%%%%%%%%%%%%%%%%%%%%%%%%%%%%%%%%%%%%%%%%%%%%%%%%%%%%%%%%%

% Requires:
% \usepackage{float}
% \usepackage{adjustbox}
% \usepackage{makecell}

\begin{table*}[t]
\caption{
Overall diagnostics on GeoBuildBench.
All runs use a maximum of 5 interaction steps.
Lower is better ($\downarrow$) for hallucinations, recovery steps, missing
objects, and failed conditions.
}
\label{tab:app-overall-diagnostics}
\centering
\small
\begin{adjustbox}{max width=\textwidth}
\begin{tabular}{lccccccc}
\toprule
Model &
Success (\%) &
\makecell{Total \\ Problems} &
\makecell{Hallucinations \\ / Prob. $\downarrow$} &
\makecell{Avg. Halluc. \\ Recovery $\downarrow$} &
\makecell{Total \\ Missing Objects $\downarrow$} &
\makecell{Total Failed \\ Conditions $\downarrow$} &
\makecell{Avg. Success \\ Steps $\downarrow$} \\
\midrule
GPT-5.1
& 78.9 & 493
& 0.401 & 1.290
& 939 & 1119
& 1.874 \\
Gemini-3-Flash
& 75.3 & 489
& 0.335 & 1.175
& 329 & 932
& 1.554 \\
Qwen3-VL-235B
& 42.2 & 491
& 2.303 & 1.743
& 2042 & 1817
& 2.304 \\
LLaMA-3.2-90B-Vision
& 21.3 & 494
& 2.381 & 1.794
& 1823 & 1584
& 2.229 \\
\bottomrule
\end{tabular}
\end{adjustbox}
\end{table*}

\section{Detailed Evaluation Diagnostics}
\label{app:analysis}

This appendix provides detailed diagnostic analyses that complement the main
results in Section~\ref{sec:benchmark-evaluation}.
While the main text focuses on high-level performance trends, the analyses here
expose \emph{how} and \emph{why} models fail, with a particular emphasis on
structural hallucinations, missing constructions, constraint violations, and
feedback utilization in the agent--environment loop.

\subsection{Overall Diagnostics}
\label{app:analysis-overall}

Table~\ref{tab:app-overall-diagnostics} summarizes overall diagnostic statistics
across all evaluated models.
In addition to success rate, we report hallucination frequency, hallucination
recovery behavior, and the total number of missing objects and failed
constraints.
As discussed in Section~5, models with higher success rates also exhibit
substantially fewer hallucinations and faster recovery from errors, indicating
more effective use of environment feedback.

\subsection{Structural Hallucination Types}
\label{app:analysis-hallucination}

Table~\ref{tab:app-hallucination-types} reports the breakdown of structural
hallucination types.
Across all models, undefined references dominate, indicating difficulty in
maintaining a consistent internal object graph during multi-step construction.
Syntax errors and output mismatches are comparatively less frequent and rarely
constitute the primary failure mode.

\begin{table*}[t]
\caption{
Breakdown of structural hallucination types (counts).
}
\label{tab:app-hallucination-types}
\centering
\small
\begin{tabular}{lccc}
\toprule
Model &
Undefined Ref. &
Syntax Error &
Output Mismatch \\
\midrule
GPT-5.1 & 87 & 33 & 74 \\
Gemini-3-Flash & 95 & 46 & 20 \\
Qwen3-VL-235B & 554 & 297 & 77 \\
LLaMA-3.2-90B-Vision & 486 & 264 & 69 \\
\bottomrule
\end{tabular}
\end{table*}

\subsection{Missing Required Objects}
\label{app:analysis-missing-objects}

Table~\ref{tab:app-missing-objects} shows the number of missing required objects
by geometric type.
Missing points and segments account for the majority of omissions, suggesting
that models often fail to explicitly instantiate all entities described in the
text, even when auxiliary constructions are allowed.

\begin{table*}[t]
\caption{
Missing required objects by object type (counts).
}
\label{tab:app-missing-objects}
\centering
\small
\begin{tabular}{lccccc}
\toprule
Model &
Points &
Segments &
Lines &
Circles &
Polygons \\
\midrule
GPT-5.1 & 406 & 286 & 128 & 35 & 84 \\
Gemini-3-Flash & 135 & 96 & 26 & 6 & 66 \\
Qwen3-VL-235B & 862 & 736 & 244 & 50 & 150 \\
LLaMA-3.2-90B-Vision & 779 & 637 & 225 & 50 & 132 \\
\bottomrule
\end{tabular}
\end{table*}

\subsection{Failed Verification Conditions}
\label{app:analysis-failed-conditions}

To keep the presentation compact, Tables~\ref{tab:app-failed-conds-gpt51}
--~\ref{tab:app-failed-conds-llama} report the top ten most frequently failed
verification conditions for each model.
Angle-value constraints, incidence relations (e.g., point-on-segment or
point-on-line), and metric constraints dominate across models, reinforcing the
observation that failures are primarily semantic rather than syntactic.

\begin{table}[H]
\caption{
Top failed verification conditions for GPT-5.1 (counts).
}
\label{tab:app-failed-conds-gpt51}
\centering
\small
\begin{tabular}{lc}
\toprule
Condition Type & Count \\
\midrule
angle\_value & 304 \\
point\_on\_circle & 135 \\
point\_on\_segment & 131 \\
point\_on\_line & 96 \\
distance\_equals & 68 \\
segment\_equality & 66 \\
perpendicular & 64 \\
parallel & 58 \\
midpoint\_of & 29 \\
angle\_bisector & 26 \\
\bottomrule
\end{tabular}
\end{table}

\begin{table}[H]
\caption{
Top failed verification conditions for Gemini-3-Flash (counts).
}
\label{tab:app-failed-conds-gemini}
\centering
\small
\begin{tabular}{lc}
\toprule
Condition Type & Count \\
\midrule
angle\_value & 332 \\
point\_on\_segment & 110 \\
distance\_equals & 108 \\
point\_on\_line & 86 \\
segment\_equality & 49 \\
parallel & 42 \\
perpendicular & 37 \\
midpoint\_of & 26 \\
concurrent & 23 \\
angle\_bisector & 19 \\
\bottomrule
\end{tabular}
\end{table}

\begin{table}[H]
\caption{
Top failed verification conditions for Qwen3-VL-235B (counts).
}
\label{tab:app-failed-conds-qwen}
\centering
\small
\begin{tabular}{lc}
\toprule
Condition Type & Count \\
\midrule
angle\_value & 473 \\
distance\_equals & 247 \\
point\_on\_circle & 190 \\
point\_on\_segment & 187 \\
perpendicular & 99 \\
point\_on\_line & 91 \\
parallel & 88 \\
triangle\_valid & 85 \\
angle\_bisector & 74 \\
segment\_equality & 65 \\
\bottomrule
\end{tabular}
\end{table}

\begin{table}[H]
\caption{
Top failed verification conditions for LLaMA-3.2-90B-Vision (counts).
}
\label{tab:app-failed-conds-llama}
\centering
\small
\begin{tabular}{lc}
\toprule
Condition Type & Count \\
\midrule
angle\_value & 439 \\
point\_on\_circle & 202 \\
distance\_equals & 175 \\
point\_on\_segment & 166 \\
parallel & 88 \\
perpendicular & 79 \\
point\_on\_line & 74 \\
angle\_bisector & 63 \\
segment\_equality & 60 \\
triangle\_valid & 54 \\
\bottomrule
\end{tabular}
\end{table}

\subsection{Feedback Utilization Indicators}
\label{app:analysis-feedback}

Finally, Table~\ref{tab:app-feedback-indicators} isolates two indicators of
feedback utilization: hallucinations per problem and the average number of steps
required to recover from hallucinations.
Lower values indicate that a model not only makes fewer structural errors but
also more effectively incorporates visual and constraint-based feedback when
errors occur.

\begin{table}[H]
\caption{
Feedback utilization indicators.
}
\label{tab:app-feedback-indicators}
\centering
\small
\begin{tabular}{lcc}
\toprule
Model & \makecell{
Hallucinations \\ / Prob. $\downarrow$ }&
\makecell{ Avg. Halluc. \\ Recovery $\downarrow$ }\\
\midrule
GPT-5.1 & 0.401 & 1.290 \\
Gemini-3-Flash & 0.335 & 1.175 \\
Qwen3-VL-235B & 2.303 & 1.743 \\
LLaMA-3.2-90B-Vision & 2.381 & 1.794 \\
\bottomrule
\end{tabular}
\end{table}
%%%%%%%%%%%%%%%%%%%%%%%%%%%%%%%%%%%%%%%%%%%%%%%%%%%%%%%%%%%%

% \begin{table}[t]
% \centering
% \small
% \begin{tabular}{llccccc}
% \toprule
% Model & Setting & Success Rate & Avg. Steps & Missing Objects & Failed Conditions & Hallucinations \\
% \midrule
% \multirow{2}{*}{Qwen3-vl-235b-a22b-instruct}
% & with vision & 42.2\% & 2.30 & 2042 & 1817 & 928 \\
% & no vision   & 39.1\% & 2.08 & 131  & 100  & 85  \\
% \midrule
% \multirow{2}{*}{Llama-3.2-90b-vision-instruct}
% & with vision & 21.3\% & 2.23 & 1823 & 1584 & 819 \\
% & no vision   & 23.1\% & 2.19 & 2408 & 2165 & 1108 \\
% \bottomrule
% \end{tabular}

% \label{tab:vision_ablation}
% \end{table}

% Additional evaluation results moved to the appendix for conciseness
\section{Additional Evaluation Results}
\label{app:additional-evaluation}

\paragraph{Cross-Lingual Evaluation.}
To assess the cross-lingual robustness of GeoBuildBench, we translate each problem into English and evaluate models under the same interactive protocol.  Table~\ref{tab:english-results} reports the success rate, average number of interaction steps, hallucinations per problem, and the total counts of missing required objects and failed constraints for GPT-5.1 and Qwen3-VL-235B on the English translation.

\begin{table*}[t]
\centering
\small
\caption{Performance on the English translation of GeoBuildBench.  Lower is better ($\downarrow$) for steps, hallucinations per problem, missing objects, and failed constraints.}
\label{tab:english-results}
\begin{tabular}{lccccc}
\toprule
Model & Success Rate (\%) & Avg. Steps $\downarrow$ & \makecell{Hallucinations \\ / Prob. $\downarrow$} & \makecell{Missing \\ Objects $\downarrow$} & \makecell{Failed \\ Constraints $\downarrow$} \\
\midrule
GPT-5.1 & 80.78 & 2.45 & 0.35 & 1509 & 1384 \\
Qwen3-VL-235B & 57.87 & 3.22 & 1.26 & 761 & 1227 \\
\bottomrule
\end{tabular}
\end{table*}

\paragraph{Difficulty-Level Analysis.}
We also examine how success varies with problem difficulty.  Table~\ref{tab:difficulty-analysis} lists the number of successful problems and success rates across four difficulty levels for each evaluated model.

\begin{table*}[t]
\centering
\small
\caption{Success counts and rates by difficulty level.  Each cell shows the number of successful problems over the total number at the given difficulty level, with the success rate in parentheses.}
\label{tab:difficulty-analysis}
\begin{tabular}{lcccc}
\toprule
Difficulty & GPT-5.1 & Gemini-3-Flash & Qwen3-VL-235B & LLaMA-3.2-90B-Vision \\
\midrule
Level 1 & $45/54$ (83.3\%) & $41/54$ (75.9\%) & $34/54$ (63.0\%) & $25/54$ (46.3\%) \\
Level 2 & $194/238$ (81.5\%) & $188/238$ (79.0\%) & $110/238$ (46.2\%) & $56/238$ (23.5\%) \\
Level 3 & $124/162$ (76.5\%) & $117/162$ (72.2\%) & $56/162$ (34.6\%) & $22/162$ (13.6\%) \\
Level 4 & $23/35$ (65.7\%) & $22/35$ (62.9\%) & $7/35$ (20.0\%) & $3/35$ (5.7\%) \\
\bottomrule
\end{tabular}
\end{table*}

\end{document}